# KLIF: An optimized spiking neuron unit for tuning surrogate gradient slope and membrane potential

Chunming Jiang, Yilei Zhang

## Abstract

Spiking neural networks (SNNs) have attracted much attention due to their ability to process temporal information, low power consumption, and higher biological plausibility. However, it is still challenging to develop efficient and high-performing learning algorithms for SNNs. Methods like artificial neural network (ANN)-to-SNN conversion can transform ANNs to SNNs with slight performance loss, but it needs a long simulation to approximate the rate coding. Directly training SNN by spike-based backpropagation (BP) such as surrogate gradient approximation is more flexible. Yet now, the performance of SNNs is not competitive compared with ANNs. In this paper, we propose a novel k-based leaky Integrate-and-Fire (KLIF) neuron model to improve the learning ability of SNNs. Compared with the popular leaky integrate-and-fire (LIF) model, KLIF adds a learnable scaling factor to dynamically update the slope and width of the surrogate gradient curve during training and incorporates a ReLU activation function that selectively delivers membrane potential to spike firing and resetting. The proposed spiking unit is evaluated on both static MNIST, Fashion-MNIST, CIFAR-10 datasets, as well as neuromorphic N-MNIST, CIFAR10-DVS, and DVS128-Gesture datasets. Experiments indicate that KLIF performs much better than LIF without introducing additional computational cost and achieves state-of-the-art performance on these datasets with few time steps. Also, KLIF is believed to be more biological plausible than LIF. The good performance of KLIF can make it completely replace the role of LIF in SNN for various tasks.

Manuscript submitted May 3, 2022. This work was supported in part by the university of Canterbury. (Corresponding author: Yilei Zhang.)

Chunming Jiang is with the department of mechanical engineering, University of Canterbury, Christchurch, 8041, New Zealand (email: cji39@uclive.ac.nz).

Yilei Zhang is with the department of mechanical engineering, University of Canterbury, Christchurch, 8041, New Zealand (email: yilei.zhang@canterbury.ac.nz).

*Index Terms*—Spiking neural network, adaptive surrogate gradient, KLIF, ReLU.

## 1. Introduction

Artificial neural networks (ANNs) have achieved remarkable success in many domains in recent years. Record accuracy at tasks such as image recognition[1-3], image segmentation[4], and language translation[5] has been achieved. However, their success is highly dependent on high-precision digital conversion[6], which requires large amounts of energy and memory. Therefore, deploying conventional ANNs on embedded platforms with limited energy and memory is still challenging.

Spiking neural networks (SNNs), regarded as the third generation of neural networks, were inspired by the biological neural system, and they mimic how information is transmitted in the human brain[7]. Unlike conventional ANNs, spiking neurons communicate and compute through discrete-time sparse events rather than continuous values. Due to being event-driven, SNNs are more efficient in terms of energy and memory consumption on embedded platforms. So far, SNNs have been used for kinds of tasks, such as image[8] and voice recognition[9].

One of the challenges in SNNs is how to train and optimize the network parameters. Currently, the existing training methods of SNNs can be classified into three types: (1) unsupervised learning, (2) indirect supervised learning, (3) direct supervised learning. The first one is inspired by the weight modification of synapses between biological neurons. A classic example is the spike time-dependent plasticity (STDP)[10-12]. Since it relies mainly on local neuronal activity rather than global supervision, STDP-based unsupervised algorithms have been limited to shallow SNNs with ⩽ 5 layers, yielding accuracy significantly lower than those provided by ANNs on complex datasets such as CIFAR-10[13-15].

The second approach is to train an ANN model firstly and then convert it to SNN with the same network structure, where the firing rate of the SNN neurons can be approximated as the analog output of the ANN neurons. For image recognition tasks, ANN-to-SNN conversion has led to state-of-the-art (SOTA) SNNs that perform close or even better than the conventional ANNs[16]. However, SNNs converted from ANNs still require a lot of inference time (about several thousand time steps) and a large amount of memory, leading to increased latency and decreased energy efficiency, which diminishes the benefits of spiking models[6, 16, 17]. The last SNN training technique is direct supervised learning, which adopts mainly the same gradient descent algorithm as in ANNs. Spikeprop pioneered the gradient descent method to train multilayer SNNs[18, 19]. It assumes that each neuron fires once in a given time window to encode the input signal and minimizes the difference between the network output and desired signal by calculating the gradient associated with these firing times.

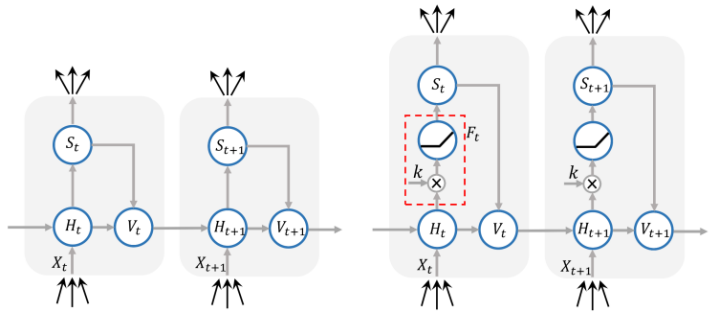

Fig. 1. Structure of spiking neurons. (a) leaky Integrate-and-Fire (LIF) model. (b) $k$-based leaky Integrate-and-Fire (KLIF) model. The dotted box represents the dynamic of function $F_t$. It incorporates a scaling factor $k$ and an activation function ReLU.

Nevertheless, the use of only a single spike per neuron has its limitations and is less suitable for processing temporal stimuli such as electroencephalography (EEG) signals, speech, or video[20]. Other subsequent works like Tempotron[21], ReSuMe[22], and SPAN[23] can utilize multiple spikes, but they can only be applied to single-layer networks. An approach treated the membrane potential as a differentiable signal to solve the problem of non-differentiation of spikes and proposed a straightforward BP algorithm to train deep SNNs with multiple spikes[24]. Recently, Wu et al. proposed a spatiotemporal backpropagation training framework for SNNs, introducing a differentiable surrogate function to approximate the derivative of spiking activity[25-27]. This method combines the spatial and temporal domains in the training phase and has yielded the best results for deep convolutional SNNs in small-scale image recognition datasets such as digit classification on the MNIST. However, for large-scale tasks, it has not been able to outperform the conversion-based approach or ANNs in terms of accuracy[16].

To further improve the performance of SNNs and decrease gaps between ANNs and SNNs, Wu et al. proposed neuron normalization. This mechanism can balance the firing rate of each neuron and avoid the loss of important information. Cheng et al. added the lateral connections between neighboring neurons and obtained better accuracy[28]. Some researchers have revised the neuron model's parameters to improve the accuracy. For example, the learnable membrane time constants in Leaky Integrate-and-Fire (LIF) neurons were utilized to make the charging and leakage process more flexible[29, 30], and an adaptive threshold spiking neuron model was proposed to enhance the learning capabilities of SNNs[31].

In this paper, we propose a novel spiking neural unit KLIF to replace the commonly adopted LIF model in SNNs. KLIF adds a learnable scaling factor that dynamically updates the slope and width of the surrogate gradient curve during training and accelerates the convergence. It also incorporates a ReLU activation function that selectively delivers membrane potential to spike firing and resetting. We verified our model on both classic static MNIST, Fashion-MNIST, CIFAR-10 datasets widely used in ANNs as benchmarks and neuromorphic N-MNIST, CIFAR10-DVS, DVS128-Gesture datasets. Experiments show that SNN with KLIF improves the test accuracy on all datasets and outperforms the SOTA accuracy.

In summary, our main work is:

1. We propose a novel spiking neural unit KLIF which can improve accuracy of models on different visual tasks by adaptive surrogate gradient descent and potential rectification.

2. We independently analyse the impact of the learnable scaling factor and rectified function. Our experiments show that the scaling factor and the ReLU activation function can independently contribute to improving accuracy of models.

3. We improve the coding layers of SNNs, which contributes to convergence and accuracy improvement of models.

## 2. Method

In Sec. 2.1, we first briefly review the LIF model and then give the dynamic equations of KLIF. In Sec. 2.2 and Sec. 2.2-2.3, we explain the benefits that KLIF brings. Finally, network structures and coding layers used in SNN

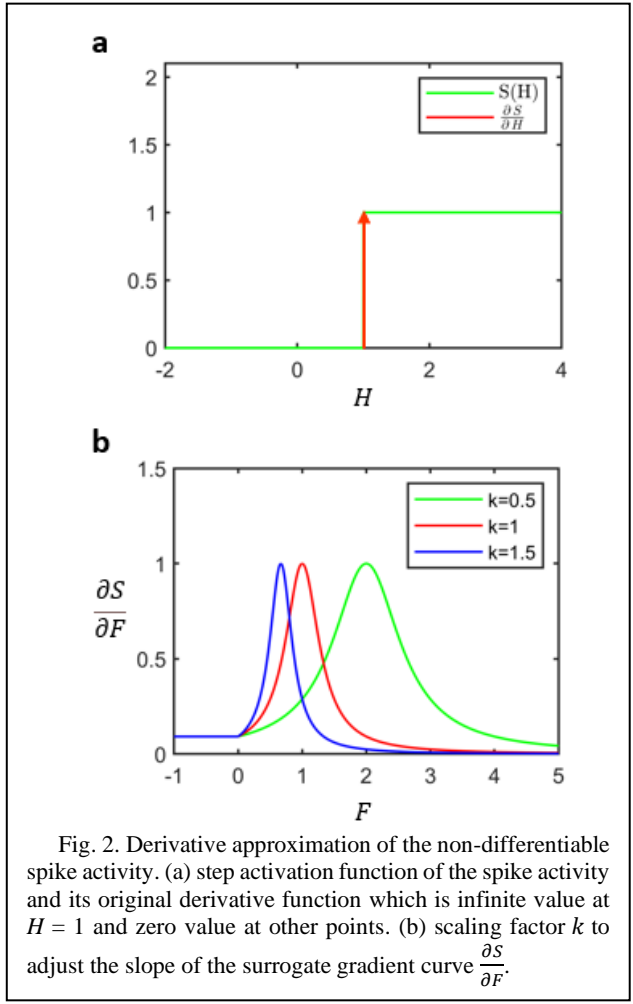

Fig. 2. Derivative approximation of the non-differentiable spike activity. (a) step activation function of the spike activity and its original derivative function which is infinite value at $H = 1$ and zero value at other points. (b) scaling factor $k$ to adjust the slope of the surrogate gradient curve $\frac{\partial S}{\partial F}$.

models, including encoding and decoding, are clarified in Sec. 2.4.

## 2.1. k-based leaky Integrate-and-Fire model

The LIF model is one of the fundamental computing units of SNNs. It is a simplified model of biological neurons and describes the non-linear relationship of input and output. The sub-threshold dynamics of the LIF spiking neuron can be modeled using equation (1).

$$\tau \frac{dV(t)}{dt} = -(V(t) - V_{\text{reset}}) + X(t) \tag{1}$$

Where $V(t)$ is the membrane potential of the neuron and $\tau$ is the time constant, $X(t)$ is defined as the weighted sum of the input spikes for each time step.

When a neuron receives inputs from the previous layer, its membrane potential will accumulate. Once the potential value exceeds the neuron's threshold, the neuron will fire a spike and promptly restore the reset potential $V_{\text{reset}}$ which is set to be 0 in this paper. To simulate the dynamic actions of LIF neurons discretely in time, we use the difference equations (2)-(4) to approximate the continuous dynamic process[29].

$$H_t = V_{t-1} + \frac{1}{\tau}(V_{\text{reset}} - V_{t-1}) + \frac{1}{\tau}X_t \tag{2}$$

$$S_t = \begin{cases} 1, & if\ H_t > V_{th} \\ 0, & otherwise \end{cases} \tag{3}$$

$$V_t = H_t(1 - S_t) + V_{\text{reset}}S_t = H_t(1 - S_t) \tag{4}$$

Where $H_t$ and $V_t$ represent the membrane potentials before and after triggering a spike at time $t$, respectively. $X_t$ denotes the external input, and $\tau$ denotes the time constant with a value of 2. $V_{th}$ denotes the firing threshold, which is 1 in this paper. $S_t$ denotes the output of a neuron at time $t$, which equals 1 if there is a spike and 0 otherwise. With (2)-(4), we describe the charging, firing, and resetting actions of the LIF neuron. Fig. 1a illustrates the dynamics of the LIF neuron.

Unlike the LIF model, we propose the $k$-based spiking neural unit (KLIF) which adds a function $F_t$ (equation (5)) between charging $H_t$ and firing $S_t$ into the LIF model (Fig. 1b). The function consists of a scaling factor $k$ and an activation function $ReLU$. As shown in Fig. 1b, a spiking neuron accumulate first its potential at time $t$, then the accumulated potential is multiplied by $k$ and passes through the ReLU function before being compared with the firing threshold. The dynamics of KLIF can be described by (2) and (5)-(7).

$$F_t = ReLU(kH_t) \tag{5}$$

$$S_t = \begin{cases} 1, & if\ F_t > V_{th} \\ 0, & otherwise \end{cases} \tag{6}$$

$$V_t = F_t(1 - S_t) + V_{\text{reset}}S_t = F_t(1 - S_t) \tag{7}$$

In section 2.3, we will discuss the reason for choosing ReLU. Compared with the LIF model, KLIF brings two benefits: adaptive surrogate gradient descent and membrane potential regulation.

## 2.2. Adaptive surrogate gradient descent

As we all know, ANNs are trained by gradient-based backpropagation (BP), which uses gradient information to

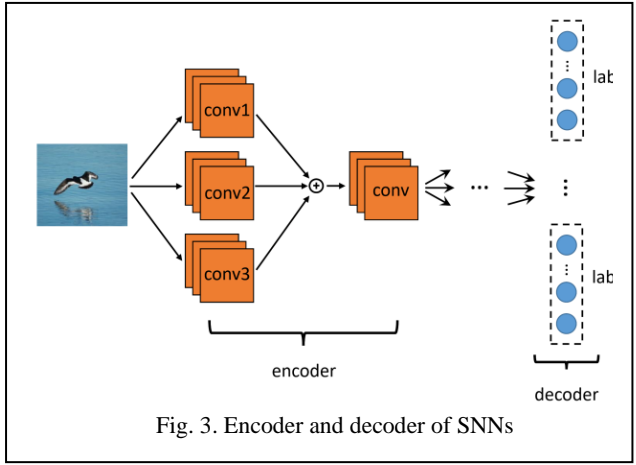
Fig. 3. Encoder and decoder of SNNs

optimize the synaptic connections and neuron parameters. Unfortunately, gradient-based optimization fails in SNNs because the firing action is non-differentiable, as described in (3). The derivative of $S_t$ is infinite at $H_t = V_{th}$, and the derivative is 0 at other places, as shown in Fig. 2a. An approach called surrogate gradient descent was proposed to address this issue[25]. The trick is to replace the derivative of the non-differentiable step function with an approximate differentiable function. It provides surrogate gradients that can be utilized to optimize the parameters of SNN efficiently during backpropagation. The differentiable function could have several forms[32]. The similarity among them is that their primitive function approximates the shape of the step function. Values of the differentiable function are relatively big around the threshold, while those away from the threshold tend to approach 0.

It has been confirmed that the type of curves of the surrogate derivative is not critical to the accuracy and the convergence speed of SNNs, but the proper curve steepness has an impact[25]. The earlier works all set the steepness empirically and do not consider the adjustment of the curve anymore during training[25]. In contrast, our work proposes the learnable scaling factor $k$ which can automatically change during the training process to fit the training data precisely.

The loss function $L$ is defined by the mean squared error (MSE). Under the principle of chain rule[33], we can calculate the gradients of the scaling factor $k^i$ in the $i$-th layer in the network according to (8).

$$\sum_t \frac{\partial L}{\partial k^i} = \sum_t \frac{\partial L}{\partial S_t^i} \frac{\partial S_t^i}{\partial F_t^i} \frac{\partial F_t^i}{\partial k^i} \tag{8}$$

$$S_t' = \frac{\partial S_t}{\partial F_t} \approx \frac{\alpha}{2(1 + (\frac{\pi}{2}\alpha(F_t - V_{th}))^2)}$$

$$= \frac{\alpha}{2(1 + (\frac{\pi}{2}\alpha(\text{ReLU}(kH_t) - V_{th}))^2)}$$

$$= \begin{cases} \dfrac{\alpha}{2\left(1 + \left(\frac{\pi}{2}\alpha V_{th}\right)^2\right)}, & H_t < 0 \\ \dfrac{\alpha}{2\left(1 + \left(\frac{\pi}{2}\alpha k\left(H_t - \frac{V_{th}}{k}\right)\right)^2\right)}, & H_t \geq 0 \end{cases}$$

(9)

In this paper, we use the derivative of arctangent $g'(x) = \frac{\alpha}{2\left(1 + \left(\frac{\pi}{2}\alpha x\right)^2\right)}$ in place of the derivative $S_t'$ of the step function in equation (9). $\alpha$ is a constant which equals 2. In equation (9), the value of the surrogate gradient $S_t'$ depends on the size of the parameter $k$. When $k$ is large, the steepness of the surrogate gradient curve is steep; conversely, it becomes flat, as shown in Fig. 2b. SNNs can adjust the gradient information by changing $k$ during training, which is more reasonable than setting it artificially. In addition, $\frac{V_{th}}{k}$ in equation (9) can be regarded as a new threshold that also depends on $k$. When k is large,

**Table 1** Network structures and training details for different datasets.

| Dataset | Simulation time | Epoch | Network structure |
|---|---|---|---|
| MNIST and Fashion-MNIST | 8 | 100 | (128C3+128C3+128C3)(encoding)-128C3-MP2-2048FC-(100FC-AP10)(decoding) |
| N-MNIST | 10 | 100 | (128C3+128C3+128C3)(encoding)-128C3-MP2-2048FC-(100FC-AP10)(decoding) |
| CIFAR-10 | 10 | 200 | (128C3+128C3+128C3)(encoding)-(256C3-256C3-256C3-MP2)*2-2048FC-(100FC-AP10)(decoding) |
| CIFAR10-DVS | 15 | 100 | (128C3+128C3+128C3)(encoding)-(128C3-MP2)*3-512FC-(100FC-AP10)(decoding) |
| DVS128-Gesture | 12 | 200 | (128C3+128C3+128C3)(encoding)-(128C3-MP2)*4-512FC- (110FC-AP10)(decoding) |

Note: nC3—Convolutional layer with n output channels, kernel size = 3 and stride = 1, MP2—2D max-pooling layer with kernel size = 2 and stride = 2, AP10—1D average-pooling layer with kernel size = 10 and stride = 10, FC—FC layer. The symbol ()*n indicates the n repeated structures.

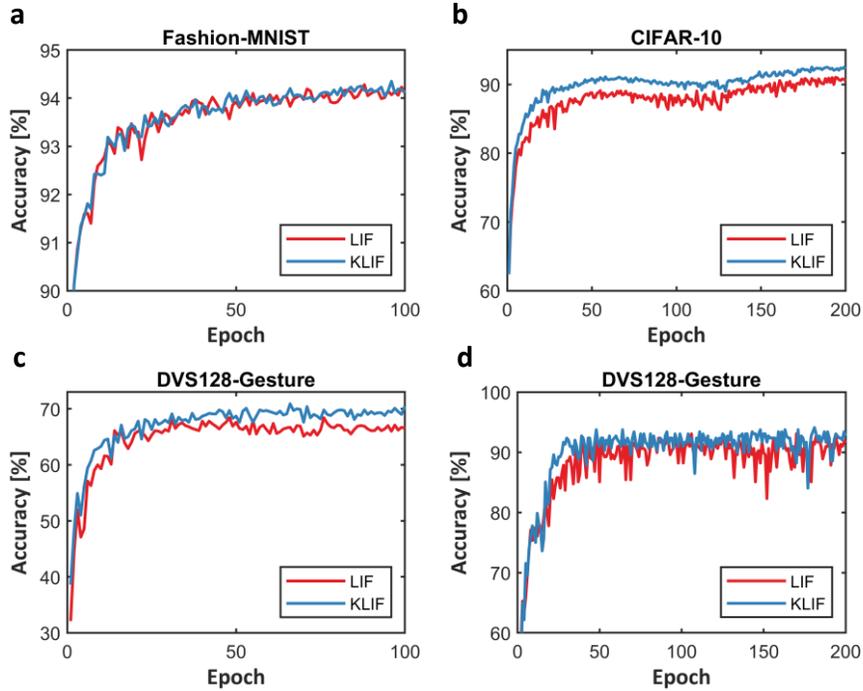

Fig. 4. The test accuracy of KLIF v.s. LIF neurons on different datasets during training.

**Table 2** Comparison between our work and the state-of-the-art methods on different datasets.

| Author | Method | SNN accuracy | | | | | |
|---|---|---|---|---|---|---|---|
| | | MNIST | Fashion-MNIST | CIFAR-10 | N-MNIST | CIFAR10-DVS | DVS128-Gesture |
| Hunsberger et al.[37] | ANN-SNN | 98.37% | - | 82.95% | - | - | - |
| Rueckauer et al.[38] | ANN-SNN | 99.44% | - | 88.82% | - | - | - |
| William et am.[17] | ANN-SNN | 99.53% | - | 88.01% | - | - | - |
| Christoph et al.[39] | ANN-SNN | - | - | 92.42% | - | - | - |
| Wu et al.[27] | Spike-based BP | - | - | 90.53 | 99.53% | 60.5% | - |
| Zhang et al.[40] | Spike-based BP | 99.62% | 90.13% | - | - | - | - |
| Lee et at.[24] | Spike-based BP | 99.59% | - | 90.95% | 99.09% | - | - |
| Shrestha et al.[41] | Spike-based BP | 99.36% | - | - | 99.2% | - | 93.64% |
| Kaiser et al.[42] | Spike-based BP | - | - | - | 96% | - | 95.54% |
| Cheng et al.[43] | Spike-based BP | 99.5% | 92.07% | - | 99.45% | - | - |
| He et al.[44] | Spike-based BP | - | - | - | 98.28% | - | 93.40% |
| Xing et al.[45] | Spike-based BP | - | - | - | - | - | 92.01% |
| Wu et al.[25] | Spike-based BP | 99.42% | - | - | 98.78% | 50.7% | - |
| Wu et al.[29] | Spike-based BP | 99.72% | 94.38% | 93.5% | 99.61% | 74.8% | 97.57% |
| Ours (with LIF) | Spike-based BP | 99.61% | 94.28% | 91.02% | 99.48% | 68.4% | 93.06% |
| Ours (with KLIF) | Spike-based BP | **99.61%** | **94.35%** | **92.52%** | **99.57%** | **70.9%** | **94.1%** |

the threshold is high; conversely, it becomes low. We use $k$ as a shared parameter with the neurons in the same layer in SNNs. This feature not only saves memory but also is biologically plausible as the neighboring neurons tend to have similar properties[29]. Notably, the parameter $k$ should be larger than 0, and cannot be too large as well, which leads to a very steep gradient curve. In practice, we find that the value of $k$ rarely becomes too small or too large as a shared parameter decided by all neurons in one layer, but just in case, we still give a boundary of it from 0.5 to 5. For the initialization, we set the values of $k$ in all layers to 1.

From one perspective, the adaptive surrogate function based on parameter $k$ makes models more flexible during training. By optimizing the value of $k$, it is possible to find the best slope and width of the surrogate function, which can speed up the convergence of the model and improve the ability to fit the training data. Since each layer has a separate $k$, which makes the surrogate function's slopes different for each layer. From another perspective, the parameter $k$ also scales the accumulated potential $H_t$ at each time step. With the increase or decrease of $k$, the potential will be amplified or reduced. It makes the charging process of neurons more controllable.

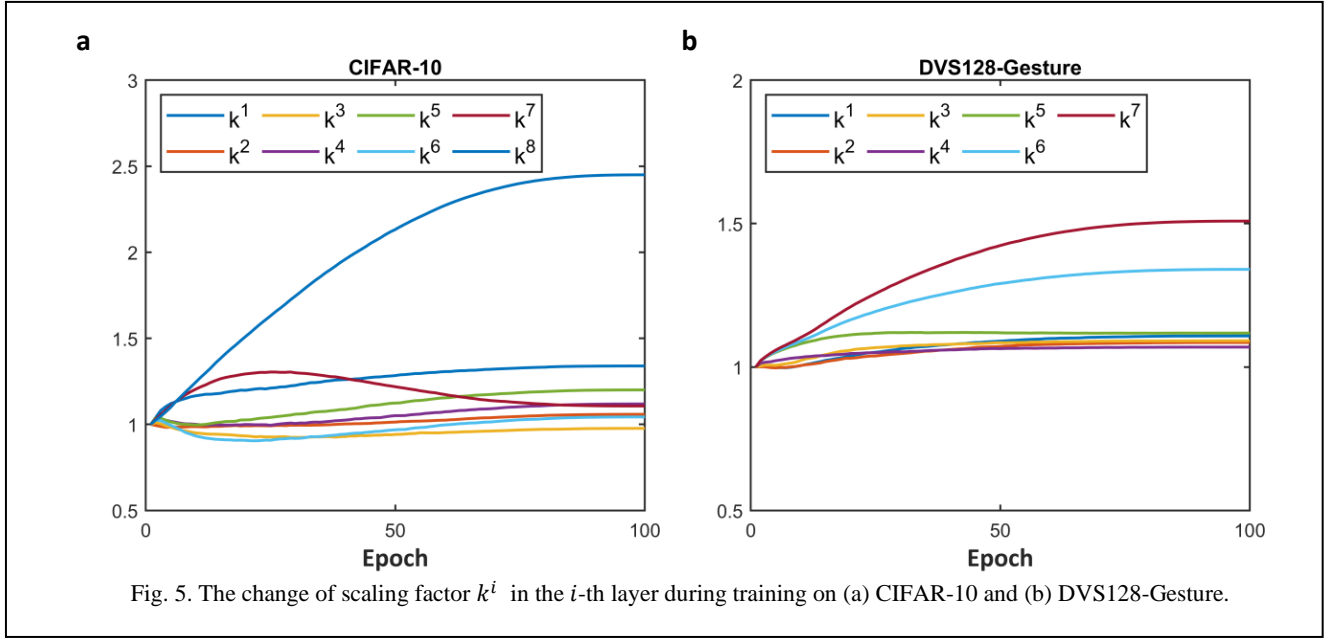

Fig. 5. The change of scaling factor $k^i$ in the $i$-th layer during training on (a) CIFAR-10 and (b) DVS128-Gesture.

### 2.3. Activation function ReLU

In addition to the scaling factor $k$, KLIF also incorporates an activation function, ReLU. ReLU keeps all the positive potentials and resets all negative potentials to 0. As $k$ is consistently less than 1, which may cause a negative potential to be an even larger negative value, it would reduce firing possibility and lead to dead neurons. ReLU limits the membrane potential from being too low to fire spikes. Also, the introduction of ReLU could save memory[34] for SNN quantitative representation when running on customed neuromorphic devices because 1) ReLU resets all negative potentials to zero, and 2) some gradients in backpropagation (BP) become zero due to ReLU. Fig. 1b shows the feedback loop of KLIF.

### 2.4. Encoding and decoding schemes

Coding layers used by SNNs are critical and decide the performance of SNNs. For the encoder, a popular method that transforms input images to spike train is rate coding. Generally, the pixel intensity of real-valued images is proportional to the firing rate in a period in rate coding. However, the conventional rate coding needs a long simulation time to present the information of images, so it is limited in training deep SNNs which have high memory requirements. An encoding layer that directly uses the first convolutional layer to encode the image information was shown to reduce the simulation time significantly and achieve a good performance[27]. Thus, we adopt a similar method to build our encoder. The difference is that we use three parallel convolutional layers rather than one convolutional layer. We add up the values of the corresponding positions of the three convolutional layers before input to spiking neurons. We adopt a voting strategy proposed in [27] for the decoder to decode the output information. It divides neurons in the output layer into several neuron populations, and each population is assigned a label. The highest determines the output class by counting the average firing rate of every population over a given time window. Fig. 3 shows the structure of the encoder and decoder.

## 3. Results and discussions

We test the proposed KLIF for classification tasks on both static datasets MNIST, Fashion-MNIST, and CIFAR-10, and neuromorphic datasets N-MNIST, CIFAR10-DVS, and DVS128-Gesture. We train SNNs by the Adam[35] optimizer with the learning rate 1e-4 and the cosine annealing[36] learning rate schedule with $T_{max} = 100$.

### 3.1. Comparison of LIF and KLIF

We compare the test accuracy of SNN models on all six datasets when using LIF and KLIF, respectively (Fig. 4). The network architectures and training details for different datasets are listed in Table 1. The hyperparameter selection like the number of filters and output feature maps are referenced in [29] which produces the best classification accuracy on different visual datasets. Except for the encoder, we use the same network architectures as those used in [29]. The batch normalization operation is used to change the input distribution after each convolutional layer. Before each fully connected layer, a dropout operation with the drop probability P = 0.5 is added to prevent overfitting. We keep the same hyperparameters and the network structure on both SNNs with different spiking neurons.

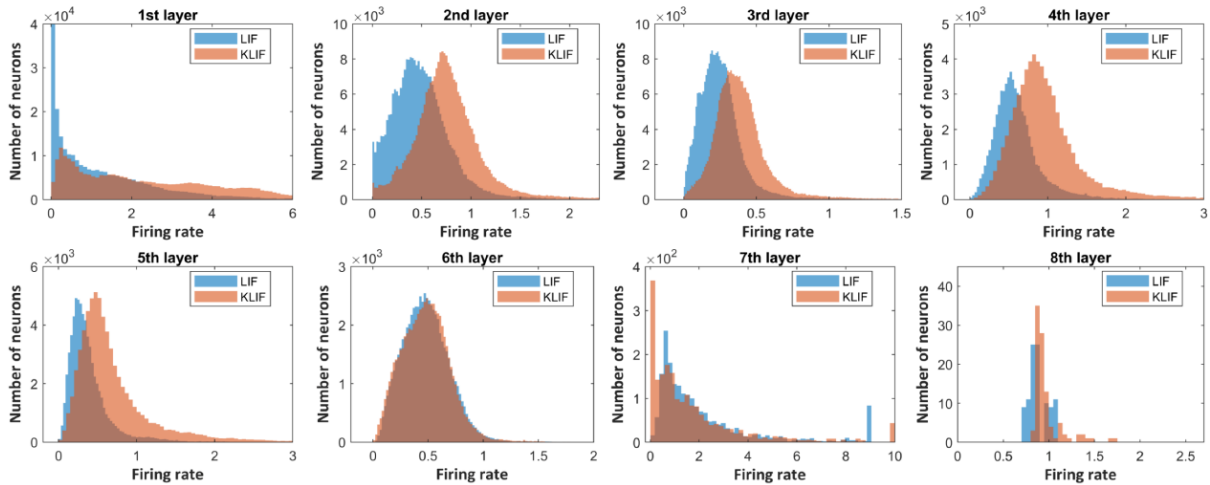

Fig. 6. The distribution of firing rate for neurons in each layer during training on CIFAR-10.

Table 3 Ablation Study of KLIF on CIFAR-10

| Neuron | VGG-16 | ResNet-18 |
|---|---|---|
| LIF | 78.45% | 85.24% |
| KLIF | **85.53%** | **89.12%** |
| Only with k | 82.96% | 87.2% |
| Only with ReLU | 81.04% | 86.66% |
| KLIF(without coding layers) | 85.53% | 89.12% |
| KLIF(with coding layers) | **86.64%** | **89.88%** |

As shown in Fig. 4, the test accuracy of the SNNs with KLIF neurons is always higher than that with LIF neurons, which verifies the validation of KLIF. In contrast, the accuracy gap is more significant on more complex datasets like CIFAR-10, CIFAR10-DVS, DVS128-Gesture than the simple Fashion-MNIST. Table 2 summarizes the results of existing state-of-the-art results. Our method achieves or approximates the best results in almost all datasets, with only no more than 10 time steps on static datasets and 15 time steps on neuromorphic datasets. Notably, accuracy listed in [29] is based on models trained for 1000 epochs, while ours are 100 or 200 epochs. With the same 100 epochs, the performance of our models is still better than that in [29] after verification.

Fig. 5 shows the change of $k$ in each layer during training on CIFAR-10 and DVS128-Gesture. It demonstrates that $k$ in each layer tends to converge during training. Fig. 6 is the distribution of firing rate comparison between the model with LIF and model with KLIF for neurons in each layer after 100 epochs training on CIFAR-10. Compared with LIF, the model with KLIF has a higher firing rate in most layers. It means that the KLIF neurons in the model are more active than LIF neurons. The result is likely to be related to the amplification of membrane potentials because most $k$ values are larger than 1 in Fig. 5. Besides, the presence of ReLU limits the lower bound of the membrane potential to 0, which makes it easier for neurons to accumulate to the threshold value in a short time and thus to fire spikes. Especially for LIF, in case the initial value of the membrane potential is at a small negative value, this makes a neuron hard to be triggered.

### 3.2. Ablation Study

In section 2, we introduce the scaling factor $k$ and ReLU function. Here we analyze their influence on models' accuracy, respectively. We selected two more commonly used network architectures for our ablation study: SNN version of VGG-16 and ResNet-18, which can demonstrate that the improved performance comes from KLIF's $k$ and ReLU rather than a certain network architecture. We trained both networks with 100 training epochs and 6 time steps under four conditions: the accuracy of LIF, only with $k$, only with ReLU, and KLIF. In VGG-16, the accuracy is 78.45%, 82.96%, 81.04%, and 85.53%, respectively. In ResNet-18, a similar conclusion can be summarized. The result clearly shows the big accuracy gap between LIF and KLIF and also indicates the benefits of $k$ and ReLU, respectively. In contrast, using $k$ alone achieves better results than using ReLU alone in both networks.

**Table 4** Accuracy of using KLIF/KLIF*

| Neuron | SNN accuracy | | | | | |
|---|---|---|---|---|---|---|
| | MNIST | Fashion-MNIST | CIFAR-10 | N-MNIST | CIFAR10-DVS | DVS128-Gesture |
| KLIF | 99.61% | 94.35% | 92.52% | 99.57% | 70.9% | 94.1% |
| KLIF* | 99.6% | 94.31% | 91.93% | 99.27% | 70.6% | 94.1% |

**Table 5** Accuracy of using KLIF with different activation functions on CIFAR-10

| Neuron | VGG-16 | ResNet-18 |
|---|---|---|
| LIF | 78.45% | 85.24% |
| KLIF(ReLU) | 85.53% | 89.12% |
| KLIF(CELU) | 84.55% | 88.7% |
| KLIF(leaky ReLU) | 85.9% | 89.11% |

Finally, the impact of the coding layers on the models' accuracy is tested. We incorporate our coding layers in both models. The test accuracy of the SNNs with our coding layers is always higher than that without our coding layers, as shown in Table 3, showing the validity of the coding layer.

### 3.3 Biological plausibility of KLIF

Biologically, neurons regulate the ion concentration difference inside and outside the membrane through the opening and closing of ion channels, thereby regulating the magnitude and range of the membrane potential. LIF neurons cannot regulate input currents and their internal potentials during training. In contrast, the parameter $k$ and ReLU regulate the magnitude and range of the membrane potential, respectively, which is more biologically plausible.

In equation (5), $F_t$ is a scaled and rectified version of $H_t$ at time t. So when computing the new $V_{t+1}$, that will be injected in equation (2) at the next time step. The scaling could also be canceled to maintain the original potential accumulation. Thus, equation (7) could be changed by dividing by $k$ on the right-hand side:

$$V_t = \frac{F_t}{k}(1 - S_t) + V_{\text{reset}} S_t = \frac{F_t}{k}(1 - S_t) \tag{10}$$

When we use this form as the expression of KLIF*, the accuracy (see Table 4) on aforementioned datasets does not change a lot compared with using KLIF.

Similarly, the ReLU function limits membrane potentials above the resting potential, which is not biologically plausible, as biological neurons can go way below the resting potential. Therefore, we replace the ReLU function in KLIF with the CELU and leaky ReLU. These functions keep the same as the ReLU in the positive part and sets the negative membrane potential as a small negative value, which is more biological plusible. The result in Table 3 shows that the test accuracy of SNN with KLIF(CELU) and KLIF(leaky ReLU) are still better than that with LIF. While the accuracy using KLIF(CELU) is slightly lower than that using KLIF(ReLU), the accuracy using KLIF(leaky ReLU) is not worse than that using KLIF(ReLU). The results also demonstrate the robustness of KLIF to different activation functions. In a sense, KLIF is more biologically plausible than LIF, because in LIF, the potential can be infinitely negative[46], which is inconsistent with the fact that biological neurons follow, while KLIF is more biologically meaningful by limiting the bounds of the negative values to fluctuate within a certain range through the activation function.

### 4. Conclusion

For a long time, there has been a relatively big performance gap between ANNs and SNNs. Kinds of methods like ANN-to-SNN conversion and direct training with spike-based BP attempt to reduce the gap. Overall, the spike-based BP is not as good as the conversion method regarding models' accuracy. However, the conversion from ANNs is based on the rate coding and usually needs a long inference time to approximate the original accuracy of ANNs, which is not efficient. More research is currently focused on how to train high-precision SNNs directly like ANNs.

In this work, we proposed the $k$-based spiking neural unit KLIF. It incorporates the learnable scaling factor $k$ and the activation function ReLU. Our experiments show that the SNN with KLIF neurons outperforms that with LIF neurons in various visual datasets. We also verify that the scaling factor and activation function can independently contribute to improving accuracy of models. The SNN updates its learnable surrogate gradients by the scaling factor over the training. The ReLU contributes to the selective delivery of positive membrane potentials. Furthermore, our coding layers with three summed convolutional layers for SNN only needs several time steps to run, which speeds up the convergence of models and improve accuracy of models.